\documentclass{article}

\usepackage{arxiv}
\usepackage{multirow}
\usepackage[utf8]{inputenc} 
\usepackage[T1]{fontenc}    
\usepackage{hyperref}       
\usepackage{url}            
\usepackage{booktabs}       
\usepackage{amsfonts}       
\usepackage{nicefrac}       
\usepackage{microtype}      
\usepackage{lipsum}
\usepackage{graphicx}
\graphicspath{ {./images/} }

\title{An Experimental Evaluation of Transformer-based Language Models in the Biomedical Domain}

\author{
Paul Grouchy\thanks{Equal Contribution} \\
 Untether AI\\

   \And
   Shobhit Jain$^*$ \\
   Manulife\\
  
  \And
   Michael Liu$^*$ \\
   Tealbook\\
  
 \And
   Kuhan Wang$^*$ \\
   CIBC\\ 
   
 \And
   Max Tian$^*$ \\
   Adeptmind\\
  
  \And
   Nidhi Arora$^*$ \\
   Intact\\ 
   
   \And
    Hillary Ngai$^*$ \\
    University of Toronto\\ 
    
   \And
   Faiza Khan Khattak \thanks{Corresponding Author: faizakhankhattak@gmail.com }  \\
   Manulife\\
   
   \And
   Elham Dolatabadi \thanks{Corresponding Author: elham.dolatabadi@vectorinstitute.ai} \\
   Vector Institute\\
   
   \And
   Sedef Akinli Kocak \thanks{Corresponding Author: sedef.kocak@vectorinstitute.ai} \\
   Vector Institute\\
 }
\begin{document}
\maketitle
\begin{abstract}
With the growing amount of text in health data, there have been rapid advances in large pre-trained models that can be applied to a wide variety of biomedical tasks with minimal task-specific modifications. Emphasizing the cost of these models, which renders technical replication challenging, this paper summarizes experiments conducted in replicating BioBERT and further pre-training and careful fine-tuning in the biomedical domain. We also investigate the effectiveness of domain-specific and domain-agnostic pre-trained models across downstream biomedical NLP tasks. Our finding confirms that pre-trained models can be impactful in some downstream NLP tasks (QA and NER) in the biomedical domain; however, this improvement may not justify the high cost of domain-specific pre-training.
\end{abstract}


\section{Introduction}
There have been increased exploration and extension of transformer-based deep learning models for NLP \cite{yang2019xlnet,liu2019roberta,devlin2018bert,lee2020biobert} recently. Using transformer-based models, one can pre-train a deep learning model on large datasets and then easily fine-tune it to adapt to downstream NLP tasks. 
There are three factors that affect the performance of these models: (a) the size of the dataset, (b) the availability of computational resources, and (c) the expressiveness of the model architecture~\cite{liu2019roberta,yang2019xlnet}. Because of these factors, the cost and complexity of developing pre-trained models are rising quickly and limit the capability of reproducing results when sufficient resources are not available.

These language models are trained on text corpora of general domains. For example, BERT~\cite{devlin2018bert}, Bidirectional Encoder Representations from Transformers has been trained on Wikipedia and BooksCorpus. There has also been a rapid growth in NLP for the biomedical domain~\cite{khattak2019survey} and many new methods including transformer-based methods are being used for different biomedical tasks involving NLP. 

The performance of language models that have been trained on general domains are not yet fully investigated in more specific domains such as biomedical, finance or legal. Therefore, it is worth investigating if large amounts of domain-specific data may help in getting better results, or if similar results may be acquired by using  smaller-sized data.

We therefore focus on biomedical domain in order to answer the following questions: 

\begin{enumerate}
    \item \textit{Does domain-specific training improve performance compared to baseline models trained on domain-agnostic corpora?}
    \item \textit{Is it possible to obtain comparable results from a domain-specific BERT model pre-trained on  smaller-sized data?} While it is established fact that with transfer learning, a model can be trained once and be reused for several tasks but there are a few cases when the model needs to be retrained. For example, when the data is dynamic and may change over time due to the data-shift; data can belong to  a wide variety of domains, therefore the model may need to be retrained for the new domain data or even sub-domain; data can be confidential to an organization hence the model needs to be retrained for that organization-specific needs. The domain and/or organization specific datasets can be very small. Moreover, not every organization has extensive computational resources required to train large models. In such cases it is helpful to know if a small domain-specific data can be used to get comparable results.
\end{enumerate}
The rest of this paper is organized as follows. We review
the existing studies related to our work in Section 2. We explain our pre-training experiments and the results in Section 3.1 and fine-tuning experiments and results in Section 3.2. We complete our paper with discussing and conclusion in Section 4 where we discuss our results and answering our questions.

\section{Related Work}
Large-scale pre-trained language models such as BERT\cite{devlin2018bert}, GPT-2~\cite{radford2019language}, RoBERTa~\cite{liu2019roberta} and GPT-3~\cite{brown2020language} have shown to outperform state-of-the-art performance in many NLP tasks such as Named Entity Recognition (NER) and Question Answering (QA). Moreover, several studies have used transfer learning and fine-tuning of these models on English NLP tasks (e.g., ~\cite{peters2019tune, yang2019end}). These language models are trained on text corpora of general domains but recently there has been a trend of training language models on the domain-specific data. For example, financial version of BERT was introduced by Araci~\cite{araci2019finbert} where he only studied sentiment classification task. Ma {\em et al}~\cite{ma2019domain} fine-tuned BERT on legal documents coming from their proprietary corpus. BioBERT~\cite{lee2020biobert} was introduced as an extension of BERT that is further pre-trained on the domain-specific biomedical corpora including PubMed and PubMed Central (PMC).

In specialized domains like biomedical, recent work has shown that using domain specific data can provide improvement over general-domain language models~\cite{gu2020domain}.
In this regard, Wang et al.~\cite{wang2018comparison} showed that word-embeddings trained on biomedical corpora captured the semantics of medical terms better than those trained on general domain corpora, but may not generalize well to downstream biomedical NLP tasks such as biomedical information retrieval. Zhao {\em et al} ~\cite{zhao2018framework} showed that word2vec~\cite{mikolov2013efficient} trained on a smaller and in-domain medical data resulted in better performance than the word2vec  trained  on a large and general domain dataset.  Also,~\cite{zhu2017semantic} found that performance decreases after 4 million distinct words of training data based on experiments with medical data from PubMed abstracts\footnote{\url{https://www.ncbi.nlm.nih.gov/pubmed/}}. In a recent study, Gu {\em et al}~\cite{gu2020domain} introduced PubMedBERT. They pre-trained the BERT from scratch with PubMed articles and a customized vocabulary (constructed from the PubMed articles). This study indicates that a proper vocabulary helps the performance of downstream tasks in specific domains. However, training the model from scratch is extremely expensive in terms of data and computation. Researchers have specifically built adaptations of BERT that attempt to address different domain related problems but the most effective pretraining process remains an open research problem~\cite{gu2020domain}. We replicated some of the BioBERT original study results, and better tune the training of BioBERT for better understanding domain specific training.

\section{Methods and Experimental Results}\label{sec:methods}
We set BERT\textsubscript{BASE} as our baseline model. We started with BERT\textsubscript{BASE}, then pre-trained it on the PubMed abstracts data (BERT\textsubscript{base}+PM) and leveraged for evaluation of downstream tasks: Name-entity recognition (NER, relational extraction (RE) (Table~\ref{table:alltasks}) and question answering (QA) (Table \ref{table:QAtasks}). This allows us to conduct a fair comparison between domain-specific pre-training and fine-tuning. PyTorch implementation of BERT\footnote{\url{https://github.com/huggingface/pytorch-transformers}} \cite{Huggingface2019} 
was leveraged and the replication experiments were conducted based on the work by McDermott et. al ~\cite{mcdermott2019reproducibility}.
 
\subsection{Pre-training language representations in the biomedical domain}
The PubMed corpora was used for pre-training consists of paper abstract from millions of samples of biomedical text. While the original BioBERT study considers combined pre-training on PubMed, PMC, and Pubmed+PMC together, our model was pre-trained only on PubMed \cite{PubMed} to check the performance based on smaller data and meet the shared computing resources.

\begin{figure}[htbp]
\begin{center}
\includegraphics[scale=0.75]{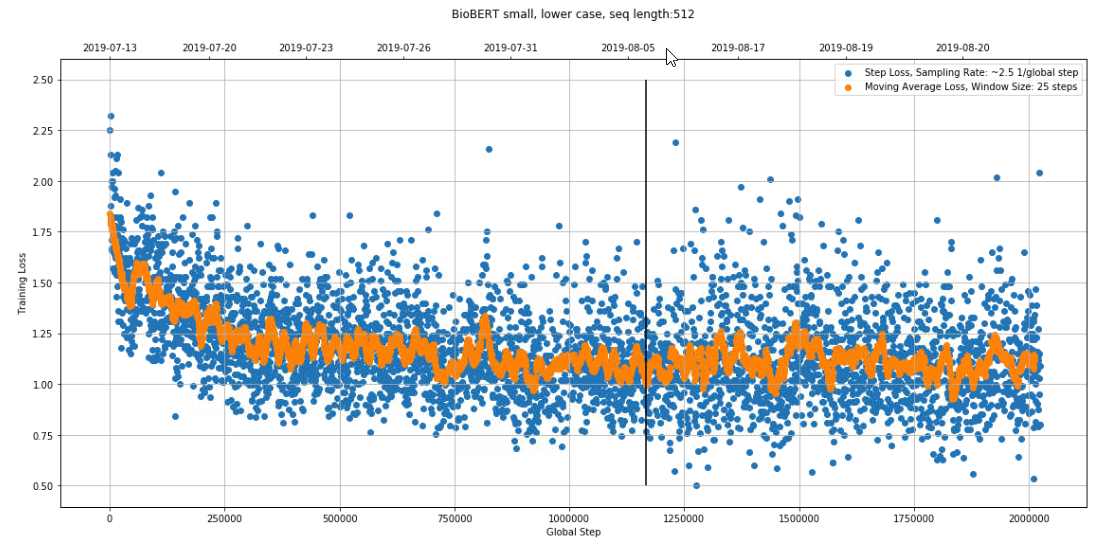}
    \caption{Training loss. The effect of chunking the data into shards is seen in the fluctuation of the training loss. The vertical line indicates separation between batch size 28 and 14 in the training paradigm. The top (bottom) x-axis indicate calendar date (global step).
    }
    \label{fig:t_loss}
\end{center}
\end{figure}

The PubMed data was processed into a format amenable for pre-training. The raw data consisted of approximately 200 million sentences in 30 GBs. The raw sentence data was batch processed into 111 chunks of ready to consume input data for BERT pre-training\footnote{This implies that the sampling of the second sentence for the Next Sentence Prediction (NSP) training task is not from the entire corpus but from within its respective chunk. As each chunk is still close to 2 million separate sentences, this was judged to be an acceptable compromise.}. Technically, the NSP task was lower-cased sentences with a maximum length of 512 and masked at the sub-token level analogous to the original BERT\textsubscript{BASE}(uncased). The original BERT models were trained on data with maximum sequence lengths of 128 for the initial 90\% of the training and 512 for the remainder. The full training loss is shown in Figure \ref{fig:t_loss}.
Generally, a longer sequence length is preferable if the corpora tends to have longer passages but the amount of data and time to train is also increased. In this experiment, the training was maintained at the maximum length of 512 tokens throughout.\\
The pre-training experiment was designed to accommodate some realistic computing resource limitations. In a time-shared computing cluster GPU resources were allocated on limited priority basis by Slurm Workload Manager. In order to facilitate stable training and to accommodate a hardware environment where resources may be reallocated to higher priority users at any time and without warning, the following training features were implemented: (1) storing model weights and gradients at regular time intervals during training; (2) querying and automating job submission from system task scheduler; (3) automatically restoring training from data chunk and step as GPU resources became available.

\subsection {Fine-tuning language representations in the biomedical domain}
Following previous work, our domain specific pre-trained model (BERT\textsubscript{base}+PM) as well as BERT\textsubscript{BASE}(uncased) were evaluated on common biomedical NLP tasks as described below.

\subsubsection{Named-entity recognition (NER)} The biomedical NER task involves the extraction of biomedical named entities such as genes, proteins, diseases and species from unstructured biomedical text. This is a challenging task because of the unique characteristics of biomedical named entities such as sharing of head nouns, spelling forms per entity, ambiguous abbreviations, descriptive naming convention, and nested names~\cite{neustein2014application}.
In the original BioBERT study nine different biomedical NER datasets were tested and their best model (BERT\textsubscript{BASE}+PM+PMC) was able to outperform state-of-the-art on the majority. In this experiment, three NER datasets including NCBI Disease~\cite{dougan2014ncbi}, JNLPBA~\cite{kim2004introduction}, and Species-800~\cite{pafilis2013species} were used. The model was trained on 50 tokens long sentences with a learning rate of $\mathit{5x10^{-5}}$. These experiments were repeated several times and F1 scores corresponding to the epoch with the minimum validation loss are reported in Table~\ref{table:alltasks}. The standard deviation in F1 scores across different runs make definitive comparisons with the results reported in BioBERT study, but a surprising outcome is that the scores for the BERT\textsubscript{BASE} starting point models are higher on average than the ones fine-tuned with our BERT\textsubscript{base}+PM. This runs counter to the notion that pre-training on biomedical domains corpora improves downstream NER.

\subsubsection{Relation Extraction (RE)} The GAD dataset~\cite{becker2004genetic} was used for the RE experiment. The goal of RE on the GAD dataset is to correctly classify whether a gene and disease are related, based on a given sentence. Two models were fine-tuned on GAD, one starting with BERT\textsubscript{BASE} and the other starting with BERT\textsubscript{base}+PM. A learning rate of $\mathit{5x10^{-5}}$ was used for all runs. The best validation F1 score over several runs for each fold is reported. Our BERT\textsubscript{base}+PM outperformed other BERT models but the differences between models are not significant (see Table~\ref{table:alltasks}).

\begin{table*}[htbp]
\centering
\normalsize
\begin{tabular}{@{}lcccccc@{}}
\toprule
\multicolumn{3}{c}{} &  \multicolumn{2}{c}{\textbf{BioBERT~\cite{lee2020biobert}}} & \multicolumn{2}{c}{\textbf{Our Experiments}} \\ \midrule
Tasks & Datasets & Metrics & BERT\textsubscript{base} & +PubMed & BERT\textsubscript{base} & BERT\textsubscript{base}+PM \\ \midrule
\multirow{3}*{NER} & NCBI Disease & F1 & 85.63 & \textbf{89.71} & 84.08$\pm$2.07 & 80.33$\pm$2.40 \\
 & JNLPBA & F1 & 74.94 & \textbf{77.49} & 76.43$\pm$2.29 & 75.16$\pm$3.63 \\
 & Species 800 & F1 & 71.63 & 74.06  & \textbf{78.38$\pm$7.87} & 74.59$\pm$5.28 \\ \midrule
RE & GAD & F1 & 79.30 & 79.83 & 79.60 & \textbf{81.50} \\ 
\bottomrule
\end{tabular}
 \caption{\label{table:alltasks}NER and RE Performance Results. Our experiments including fine-tuning of the BERT\textsubscript{BASE}(uncased) model and the BERT\textsubscript{BASE}(uncased) model further pre-trained on PubMed abstracts (+PM). For comparison, the results of BioBERT study using BERT\textsubscript{BASE} (i.e. Wiki+Books) and the BioBERT version of the PubMed trained model (+PubMed) are also shown. Best scores are shown in bold.}
\end{table*}
\vspace{0cm}

\subsubsection{Question Answering (QA)} The goal of QA tasks is to automatically find the answer to a question posed in human language, usually from a context paragraph. In this study, we explored the performance of BERT and BioBERT on two common biomedical QA tasks including BioASQ~\cite{tsatsaronis2015overview} and PubmedQA~\cite{jin2019pubmedqa}. For this task, three versions of BERT model (BERT\textsubscript{large}, BERT\textsubscript{base}, and our BERT\textsubscript{base}+PM) were fine-tuned on the QA datasets.

\textbf{BioASQ:} As was suggested in the BioBERT study~\cite{lee2020biobert}, all BERT models were initially fine-tuned on the SQuAD~\cite{rajpurkar2016squad} dataset (with intermediate evaluations), and then on the BioASQ training set before finally evaluating on the BioASQ test sets (Table~\ref{table:QAtasks}). It has been also shown in~\cite{yoon2019pretrainBioQA} that pre-training BERT on SQuAD 1.1 generated better results when fine-tuned on BioASQ in comparison to the model pre-trained on SQuAD 2.0. For tuning the model on SQuAD 1.1 and SQuAD 2.0, we took inspiration from the training schemes outlined in~\cite{liu2019roberta} and~\cite{lan2019albert} to adjust the hyperparameters, namely the learning rate, $\beta_2$, learning rate schedule, batch size, and number of training epochs. We found that training with a cosine learning rate schedule with no warm-up steps with $\beta_2 = 0.98$ consistently resulted in the best performance, and thus carried over the heuristics generated from fine-tuning on SQuAD to further fine-tune our models on BioASQ.
\textit{Training on domain-specific data:} Overall, there is some evidence (Table~\ref{table:QAtasks}) that pre-training on biomedical domain corpus improves performance on the downstream BioASQ QA task. However, the improvement is not so large as to be entirely convincing and carefully fine-tuned BERT models (such as BERT\textsubscript{large} in our case) can perform comparably to BioBERT.\\\textit{Zero-shot setting:} We also conducted an experiment in order to evaluate the performance of the BERT models on BioASQ datasets in a zero-shot setting where BERT\textsubscript{BASE} was fine-tuned on SQuAD and \emph{not} on BioASQ training data. We averaged the scores over the 5 test sets within Strict Accuracy (S), Lenient Accuracy (L), and Mean Reciprocal Rank (M) metrics and the results were 31.37, 46.8, and 37.16, respectively. As we can see the zero-shot evaluation results on test sets are worse than the results indicated in Table~\ref{table:QAtasks} which emphasizes the effects of fine-tuning on BioASQ.\\\textit{Number of epochs:} Additionally, we found that fine-tuning BERT\textsubscript{large} model on BioASQ for more epochs than the typically recommended 1-3 epochs resulted in much better results. Our best results across the three evaluation metrics on BioASQ came from fine-tuning BERT\textsubscript{large} for 20 epochs and BERT\textsubscript{base} for 4 epochs with initial learning rate of $5x10^{-6}$.\\
\begin{table*}[ht]
\centering
\normalsize
\begin{tabular}{@{}lc|cc|ccc@{}}
\toprule
\multicolumn{2}{c}{} & \multicolumn{2}{c}{\textbf{BioBERT~\cite{lee2020biobert}}} & \multicolumn{3}{c}{\textbf{Our Experiments}} \\ 
\midrule
Datasets & Metrics &  BERT\textsubscript{base} & +PubMed & BERT\textsubscript{large}$^*$ & BERT\textsubscript{base} & BERT\textsubscript{base}+PM \\ 
\midrule
\multirow{3}*{BioASQ (4b)}& S & 27.33 & 27.95& \bf 31.8 & 31.2&31.54 \\
                      & L & 44.72 & 44.10& \bf 51.8 & 44.58&48.36 \\
                      & M & 33.77 & 34.72& \bf 40.0 & 36.25&38.39 \\
\multirow{3}*{BioASQ (5b)}& S & 39.33 & 46.00  & 43.0 & 41.61& \bf46.05 \\
                      & L & 52.67 & \bf60.00  & 55.8 & 56.8&57.94 \\
                      & M & 44.27 & \bf51.64& 48.2 & 47.86& 50.54\\
\multirow{3}*{BioASQ (6b)}& S & 33.54 & \bf42.86& 35.8 & 36.55&37.57 \\
                      & L & 51.55 & 57.77& 54.4 & \bf 59.3&58.87 \\
                      & M & 40.88 & \bf48.43& 43.4 & 47.59&46.5 \\
\cmidrule{2-7}
Average               & S & 33.4 & \bf 38.93 & 36.86 & 37.48 & 38.39\\
over                  & L & 49.65 & 52.68& \bf55.12 & 53.56 & 55.06\\
4b,5b, and 6b         & M & 39.6 & \bf 44.93& 43.86 & 43.9  & 45.14\\
\midrule
\multirow{2}*{PubMedQA} & K-Fold Acc & \textemdash & \textbf{57.28} & 56.52 & 55.20 & 56.20 \\
                    & K-Fold F1  &\textemdash & \textbf{28.70} & 26.14 & 23.71 & 23.98 \\
 \bottomrule
\end{tabular}
\caption{\label{table:QAtasks}
The performance results of three versions of BERT model (BERT\textsubscript{large}, BERT\textsubscript{base} and BERT\textsubscript{base}+PM) on two distinct QA datasets. For comparison, the results of BioBERT study using BERT\textsubscript{BASE} (i.e. Wiki+Books) and the BioBERT version of the PubMed trained model (+PubMed) are also shown. Since, we already explored latest versions of BERT\textsubscript{large} and BERT\textsubscript{base}, we didn't examine the BERT\textsubscript{base} version used in the BioBERT study. Best scores are shown in bold. *Used BERT$_{large}$ as other versions were not providing  good results.}
\end{table*}

\textbf{PubMedQA:} 
All three versions of the BERT models were fine-tuned on PQA-L i.e., 1k expert-annotated generated QA instances of the PubMedQA dataset with the results shown in Table~\ref{table:QAtasks}. Since PubMedQA hasn’t been experimented in the BioBERT study, in order to make a fair comparison between our study and the BioBERT study~\cite{lee2020biobert}, we ran an additional experiment where we fine-tuned  PubMed trained model (+PubMed) from the BioBERT study on the PubMedQA. 10-fold cross-validation was performed with only 450 training instances in each fold of validation. As seen in Table~\ref{table:QAtasks}, the BioBERT version of the PubMed trained model (+PubMed) has the highest accuracy 57.28 and F1 Score 28.27. This suggests that pre-training on biomedical corpus improves the performance of the downstream PubMedQA task. However, the performance improvement compared to BERT\textsubscript{large}, BERT\textsubscript{base}, and BERT\textsubscript{base}+PM is minimal.

\subsubsection{Text Summarization}
Text summarization refers to automatic generation of summary of a given text. Extractive summarization is done by extracting the most important sentences from the document that summarize the whole document. Abstractive summarization refers to condensing the document into shorter versions while  preserving its meaning \cite{gupta2010survey}. 

\begin{table*}[hbt!]
\centering
\normalsize
\begin{tabular}{ llll|llc|lll } 
 \hline
&\multicolumn{3}{c}{\textbf{CNN/DailyMail}} &\multicolumn{3}{c}{\textbf{XSum}} &\multicolumn{3}{c}{\textbf{BioASQ 7b \tiny{(Our experiments)}}}\\

$\#$ docs (train/val/test) &\multicolumn{3}{c}{\small{196,961/ /12,148/10,397}} &\multicolumn{3}{c}{\small{204,045/11,332/11,334}} &\multicolumn{3}{c}{\small{22,462/4,814/4,800}}\\
 \hline
 & \scriptsize{ROUGE-1} &  \scriptsize{ROUGE-2} &  \scriptsize{ROUGE-3}
  &  \scriptsize{ROUGE-1} &  \scriptsize{ROUGE-2} &  \scriptsize{ROUGE-3}
   &  \scriptsize{ROUGE-1} &  \scriptsize{ROUGE-2} &  \scriptsize{ROUGE-3}\\
 \hline
\textbf{Abs} &\textbf{41.72}& 19.39 & \textbf{38.76}  & 38.81 &16.33 &31.15 & 38.20 & \textbf{27.12}  & 34.15 \\ 
 
\textbf{ExtAbs} & \textbf{42.13}  & 19.6 &\textbf{39.18} & 38.76& 16.50 & 31.27  & 38.89  & \textbf{29.29} & 36.01\\ 
 
\textbf{Ext} & \textbf{43.25}  & 20.24&\textbf{39.63} & $\dagger$  &$\dagger$ & $\dagger$ & 33.30 &\textbf{26.50}  & 32.20\\ 
\bottomrule
\end{tabular}
\caption{\label{table:TextSum} Text summarization F1 score. $\dagger$ Results were poor hence were not reported by the authors \cite{liu2019text} }
\end{table*}

BERTSum \cite{liu2019fine,liu2019text} is a simple variant of BERT, for text summarization that produces abstractive and extractive summaries. BERTSumAbs is based on an encoder-decoder architecture where the encoder is pre-trained whereas decoder is a randomly initialized transformer which is trained from scratch. There are three versions of BERTSum. BERTSumAbs produces abstractive summary, BERTSumExt outputs extractive summary, and  {BERTSumExtAbs also produces abstractive summary but is based on two-stage approach where the encoder is fine-tuned twice, first with an extractive objective followed by an abstractive one.

 We explored the performance of these methods on BioASQ 7b\footnote{\url{http://bioasq.org/participate/challenges_year_7}} dataset consisting of links to research papers as well as summaries are embedded in the json file itself. BERT was trained from scratch on BioASQ 7b. The results are shared in Table \ref{table:TextSum}. 
 For CNN/DailyMail dataset all three methods outperform on BioASQ dataset according to ROUGE-2 score, while produce comparable results in all other cases. For XSum dataset \cite{narayan2018don}, we have comparable results according to ROUGE-1, while BioASQ 7b performs better according to ROUGE-2 and ROUGE-3 (for Ext summary results were not reported by the authors \cite{liu2019text}). This maybe due to the fact that BioASQ dataset is much smaller in size as compared to the other two datasets. This also shows that smaller-sized dataset can be used for text-summarization.
 \section{Discussion and Conclusion}\label{sec:discussion}
In this study, we present our findings for experiments conducted in conjunction with further pre-training of BERT model in the biomedical domain as well as evaluating both the domain specific and domain agnostic pre-trained models across downstream biomedical NLP tasks. Our experiments also included considering data and optimization related factors. Further pre-training was conducted on PubMed abstracts only to evaluate the performance of the models trained on smaller-size datasets and also the effects of learning rate was carefully evaluated for fine-tuning tasks. 

\textit{(1) Does domain-specific training improves performance?} This study confirms that unsupervised pre-training in general could improve the performance on fine-tuning tasks. However, the effectiveness of \textit{domain specific} pre-training as a way of \textit{further} improving the performance of supervised downstream tasks does not significantly outperform the effectiveness of domain agnostic pre-trained models considering the high cost of domain-specific pre-training which makes it challenging for most of researchers and NLP developers. In the biomedical domain, however, this conclusion may not be wholly substantiated owing to a lack of consistent evidence particularly in downstream NER, QA, and Text Summarization tasks. For the SQuAD task, it has been shown that fine-tuning results depend on the size and duration of training \cite{Talmor2019} \cite{WieseWN17}. However, given the small size of the biomedical QA (BioASQ) datasets, it is not possible to run the same experiments, as the current pre-trained models easily overfit. Therefore, it should be beneficial for the biomedical community to curate and expand the magnitude of benchmark datasets. Understandably it is difficult and expensive, but with the ever increasing size of the deep-learning models and significant advances in the development of pre-training language representations, it is necessary in order to facilitate reproducible research for health.

\textit{(2) Is it possible to obtain comparable results using BERT model pre-trained  on smaller-sized data?} We  present results of the experiments that were conducted using the models pre-trained on the PubMed abstracts only. These results were comparable to the results produced by the model trained on the PubMed, PMC, and Pubmed+PMC together (please check ~\cite{lee2020biobert}). Although it requires further investigation but empirically it shows that small-sized datasets may be used as a surrogate. Using small-sized dataset can be especially useful when the models need to retrained instead of using pre-trained publicly available models due to  (a) data-shift, (b) wide variety of data-domains, (c) confidential data not publicly available to train the model on, (d) small-size of the domain specific data available, and (e) a lack of computing resources.

We position these experiments as complementary to existing literature on the applications of transformer-based 
models for biomedical NLP. Our results provide some evidence for the validity and the limitations of existing language representations for pre-training $\&$ fine-tuning in biomedical domain.

We would like to expand our experiments on other variants of BERT, with more domain specific datasets and a variety of downstream tasks. This would also allow us to use our experiments for other medical corpora\footnote{\url{http://www.nactem.ac.uk/resources.php}} for external validation and generalization of the results.

\section*{Acknowledgements}
We want to thank Vector Institute industry sponsors, researchers and technical staff who participated in the Vector Institute's NLP Project (\url{https://vectorinstitute.ai/wp-content/uploads/2020/12/nlp-report-final.pdf}).

\typeout{}
\bibliographystyle{unsrt}  
\bibliography{references}

\end{document}